\pdfoutput=1

\documentclass[10pt,twocolumn,letterpaper]{article}
\usepackage{comment}
\usepackage[pagenumbers]{cvpr}

\usepackage{xspace}
\usepackage{booktabs}
\newcommand{\fw}{R2R\xspace}

\usepackage{multirow}

\definecolor{cvprblue}{rgb}{0.21,0.49,0.74}
\usepackage[pagebackref,breaklinks,colorlinks,allcolors=cvprblue]{hyperref}

\def\confName{CVPR}
\def\confYear{2027}
\let\oldtwocolumn\twocolumn
\renewcommand\twocolumn[1][]{%
    \oldtwocolumn[{#1}{%
        \centering
        \vspace{-10pt}
        \includegraphics[width=\textwidth]{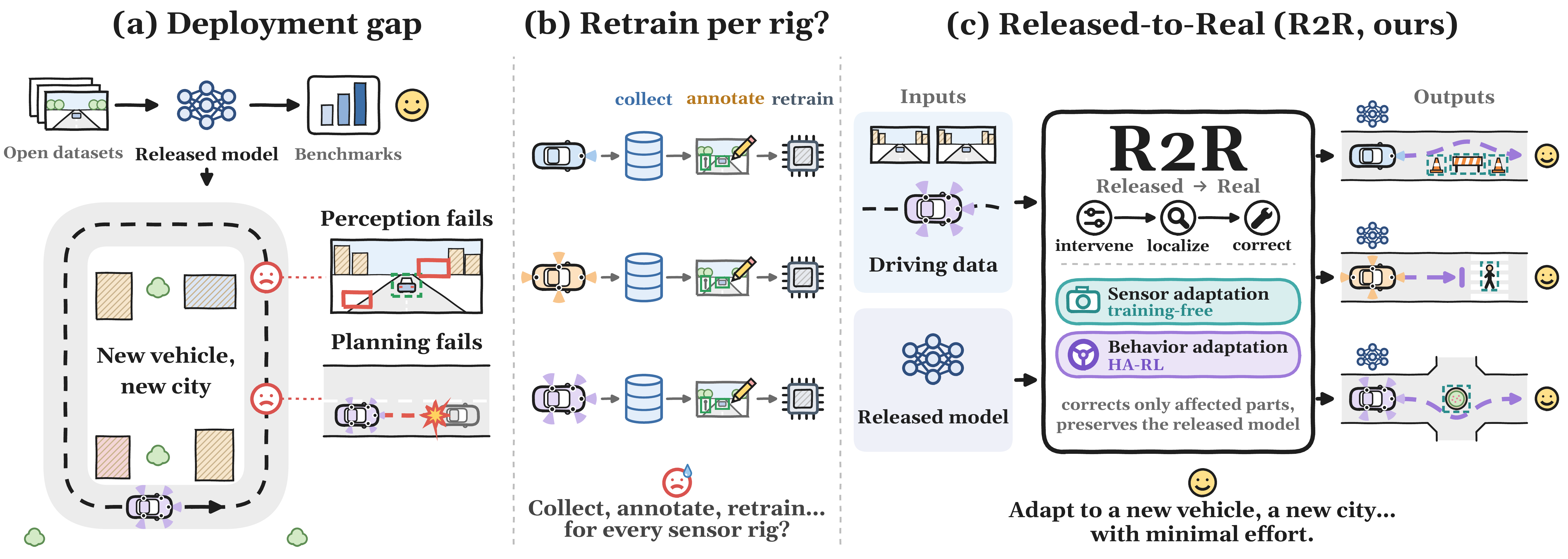}
        \vspace{-6pt}
        \captionof{figure}{\textbf{Released-to-Real (R2R).} (a) A released driving model that scores well on benchmarks fails in perception and in planning on a new vehicle in a new city. (b) Retraining would require collecting and annotating data for every sensor rig. (c) R2R uses ordinary driving data from the target vehicle to localize what remains coupled to the source system and corrects only that part: sensor adaptation with frozen weights and behavior adaptation with Human-Anchored Reinforcement Learning (HA-RL).}
        \label{fig:teaser}
        \vspace{6pt}
    }]
}

\title{Beyond the Benchmark: Localizing Released Autonomous Driving Models to Real Vehicles}

\author{Morui Zhu$^{1,2}$ \qquad Deyuan Qu$^{1}$ \qquad Qi Chen$^{1}$ \qquad Qing Yang$^{2}$\\[4pt]
{\normalsize $^{1}$Toyota North America R\&D \qquad $^{2}$University of North Texas}
}

\begin{document}

\maketitle
\begin{abstract}
Released autonomous-driving models achieve strong benchmark performance, yet deploying them on a new real vehicle remains challenging. We study this deployment gap and find that released models can remain coupled to assumptions from their source sensing, data, and execution processes even after conventional adaptation.
We introduce Released-to-Real (R2R), a deployment-localization methodology that identifies such source-system coupling and corrects only the affected components while preserving the released model. For data and behavior adaptation, we find that the mismatch between trajectory-level optimization and receding-horizon execution enters through the ego state rather than the observation, and introduce Human-Anchored Reinforcement Learning (HA-RL), which corrects it on recorded target driving without the costly rollouts of generative closed-loop reinforcement learning (RL). For physical sensor adaptation, R2R localizes and corrects source-dependent spatial and temporal representations with frozen perception weights, requiring neither target annotations nor
retraining. HA-RL improves the closed-loop success rate of a released diffusion-based planner from 82.2\% to 95.2\%, while sensor adaptation across multiple released detectors and three real sensor rigs improves the vehicle mean average precision (mAP) of a released multi-camera detector from $0.041$ to $0.420$ on a five-camera target rig. These results provide a practical path for preparing released models before deployment on new real vehicles. Our project page is publicly available at \url{https://released-to-real.github.io}.
\end{abstract}
    
\section{Introduction}
\label{sec:intro}

Open datasets and released models have made strong autonomous-driving perception and planning systems increasingly accessible~\cite{navsim,diffusiondrive,sparse4dv3,streampetr}.
Yet strong benchmark performance does not make a released model ready for deployment on real vehicles (\cref{fig:teaser}a).
Differences in sensor configuration, data distribution, and execution process can invalidate assumptions learned under the source system.

Existing work addresses individual aspects of this deployment gap. Cross-dataset and cross-camera methods improve perception transfer across sensing configurations ~\cite{zheng2023crossdataset,unidrive,coin3d}, while imitation learning (IL), RL post-training, and world models improve planner adaptation and closed-loop behavior ~\cite{recogdrive,worldengine,roxdrive}.
These methods typically address transfer during model development or adapt individual components of the driving system.
They leave a different practical question largely unresolved: \emph{given an already trained released model and limited driving data from a new vehicle, how can we systematically localize it for real-vehicle deployment?}

This deployment gap persists even after seemingly necessary adaptation has been performed. On the planning side, conventional post-training of a released planner, IL followed by RL with a trajectory-level driving score, raises its open-loop score from $0.698$ to $0.809$ but reduces the closed-loop success rate from 82.2\% to 78.1\%, with the collision rate increasing from 4.4\% to 12.2\%. On the perception side, a released multi-camera 3D detector achieves only $0.041$ vehicle mAP on a five-camera target rig even with the correct target calibration. These failures reveal a common cause: released models can remain coupled to assumptions of the source system even after explicit sensor-interface alignment (SIA) with the known target calibration.

To close this gap, we introduce \textbf{R2R}, a unified deployment-localization methodology for identifying and correcting such
source-system coupling (\cref{fig:teaser}c).
R2R uses controlled intervention: it varies one deployment variable while holding the remaining inputs, representations, or model parameters fixed whenever possible, localizes the source dependency, and corrects only the affected component.
We instantiate this methodology for two recurring deployment mismatches: data and behavior adaptation and physical sensor adaptation.

For data and behavior adaptation, R2R first re-renders the target cameras as the cameras the planner was trained with, and then reuses ordinary
target-vehicle driving for supervision, behavioral grounding, and target-domain evaluation.
We identify a reward--execution mismatch: a trajectory-level objective can favor behavior that scores well over the full predicted horizon but becomes unsafe when only its initial segment is repeatedly executed.
Direct generative closed-loop RL resolves this execution inconsistency by evaluating the planner under state-dependent rollouts.
However, it regenerates the camera observations for every rollout, which makes each rollout costly and requires a world model of the target environment.
We find that the mismatch enters through the ego state: letting the ego state alone follow the planner's execution, on recorded observations, recovers the closed-loop gain.
This finding motivates \emph{HA-RL}, which combines planner-induced execution with real target-domain observations and human behavioral anchors.
HA-RL improves the closed-loop success rate of the released planner from 82.2\% to 95.2\%.

For physical sensor adaptation, R2R applies the same re-rendering to the cameras of the released
perception model, keeps the model frozen, and uses controlled probes to identify representations that remain coupled to the source sensing process.
We localize source-dependent spatial and temporal priors and correct only the affected representations using the known source and target sensor configurations, without target annotations or gradient updates.
Although the concrete corrections are architecture dependent, the same localization procedure applies across models.
Across three real sensor rigs, R2R consistently improves detectors with implicit position embeddings and with explicit anchor sampling; on the five-camera rig, the former improves from $0.041$ to $0.420$ mAP with frozen weights.

Our contributions are threefold:
\begin{itemize}
    \item We identify a practical deployment gap between strong benchmark performance and real-vehicle deployment, and introduce R2R, a unified deployment-localization methodology based on controlled intervention. R2R identifies source-system coupling and corrects only the affected
    objective, interface, or representation while preserving the remaining released model.

    \item We show that trajectory-level optimization can conflict with receding-horizon execution in planner adaptation, and that this mismatch enters through the ego state rather than the observation. We introduce HA-RL, which corrects it on recorded target driving by combining planner-induced execution with
    human behavioral anchors. HA-RL matches generative closed-loop RL in closed-loop driving at a 19$\times$ lower rollout cost, improving the closed-loop success rate from 82.2\% to 95.2\%.

    \item We show that released perception models can retain source-sensor spatial and temporal priors even after the correct target calibration is supplied. We introduce a training-free localization procedure that identifies and corrects these sensor-dependent representations with frozen perception weights, improving transfer across public datasets and three real sensor rigs.
\end{itemize}

\section{Related Work}
\label{sec:related}

\textbf{Post-training planners and the open-loop / closed-loop gap.}
End-to-end planners are commonly post-trained with IL
and RL using open-loop benchmark objectives
~\cite{recogdrive,autovla,diffusiondrivev2,navsim}.
Closed-loop studies in photorealistic environments show that open-loop
gains do not reliably transfer to closed-loop safety and that
open-loop reinforcement fine-tuning can trade safety margin for
progress~\cite{bao2026navsafe}.
This discrepancy has been attributed to observational shift and
objective mismatch~\cite{zhao2026bridgesim}, and score gains under
optimization can reverse when the execution interface changes
~\cite{zhu2026score}.
Recent methods improve closed-loop behavior through additional
post-training or generative environments
~\cite{rad,adr1,roxdrive,nord}, but are primarily developed and
evaluated on public datasets and simulators.
Pseudo-simulation~\cite{navsimv2} evaluates a planner from observations synthesized around the recorded trajectory; HA-RL instead feeds the executed ego state back to the planner on the recorded observations and uses the resulting rollout as a training reward.
These studies establish the importance of closed-loop optimization,
but leave open how an already released planner should be localized to
the data and execution process of a new real vehicle.
R2R addresses this released-model deployment setting using limited
driving data collected from the target vehicle.

\textbf{Camera 3D detection across sensor rigs.}
Camera 3D detectors can lose accuracy when the camera configuration
changes.
Existing approaches improve cross-rig transfer through augmented or
re-rendered views, intrinsics-decoupled depth, domain alignment, or
configuration-aware training
~\cite{dgbev,pdbev,udga,unidrive,coin3d}.
Closest to our setting, Zheng~\etal~\cite{zheng2023crossdataset} align
the target coordinate origin with the source ego height, with the
alignment selected using labeled target data.
Training-free analyses of vision foundation models further show that
learned features can contain content-independent positional components
~\cite{insid3}.
However, existing approaches generally either modify the training
process, require target supervision, or study representation bias
outside the deployment of released driving detectors.
R2R instead considers how a released detector can be
localized to a new physical sensor rig without target annotations or
retraining.

\section{R2R Deployment-Localization Methodology}
\label{sec:r2r}

R2R treats the deployment of a released model on a new real vehicle as
a localization problem: the model may remain coupled to assumptions of
the source system that no longer hold on the target vehicle.
Rather than retraining the entire model (\cref{fig:teaser}b), R2R identifies such
source-system coupling through controlled intervention, corrects only
the affected component, and validates the localized model under the
target deployment process.

\subsection{Problem Setup}
\label{sec:problem_setup}
Let $f_{\Theta}=(q_{\psi},p_{\theta})$ denote a released driving model, where $q_{\psi}$ and $p_{\theta}$ are the perception and planner modules, respectively. 
The model is pretrained on an open source dataset $\mathcal{D}_s$ collected under sensor configuration $\Gamma_s$ and developed under a specific execution and evaluation process $\mathcal{E}_s$. 
The target vehicle provides a finite human-driving dataset $\mathcal{D}_t$ and operates with sensor configuration $\Gamma_t$ and execution process
$\mathcal{E}_t$. 
We denote the source and target systems as $\mathcal{S}_s=(\mathcal{D}_s,\Gamma_s,\mathcal{E}_s)$ and $\mathcal{S}_t=(\mathcal{D}_t,\Gamma_t,\mathcal{E}_t)$.
Our goal is to localize the released model $f_{\Theta}$ to $\mathcal{S}_t$ using $\Gamma_s$, $\Gamma_t$,  $\mathcal{D}_t$, and open road maps, without manually annotated objects in $\mathcal{D}_t$.

The key challenge lies in changing the explicit deployment interface does not necessarily remove assumptions learned in $\mathcal{S}_s$.
We refer to these residual dependencies as \emph{source-system coupling}.
R2R provides a unified methodology for identifying and correcting such coupling through controlled intervention.
We vary one deployment variable while holding the remaining inputs or model parameters fixed, thereby isolating the source assumption that fails to transfer.
R2R then localizes the resulting source-system coupling, corrects only the affected objective, interface, or representation, and validates the corrected model under $\mathcal{S}_t$.

In practice, the deployment gap can arise from the differences in $\mathcal{D}_t$, $\mathcal{E}_t$, and $\Gamma_t$ from their source counterparts.
We therefore instantiate R2R along two complementary adaptation paths that share one first step, the alignment of the sensor interface to $\Gamma_s$: (1) \emph{data and behavior adaptation}, which then addresses changes in $\mathcal{D}_t$ and $\mathcal{E}_t$, and (2) \emph{physical sensor adaptation}, which then addresses what the released perception model retains of $\Gamma_s$ while keeping it fixed.

\subsection{Localized Corrections}
\label{sec:behavior_adaptation}

\begin{figure*}[t]
  \centering
  \includegraphics[width=\textwidth]{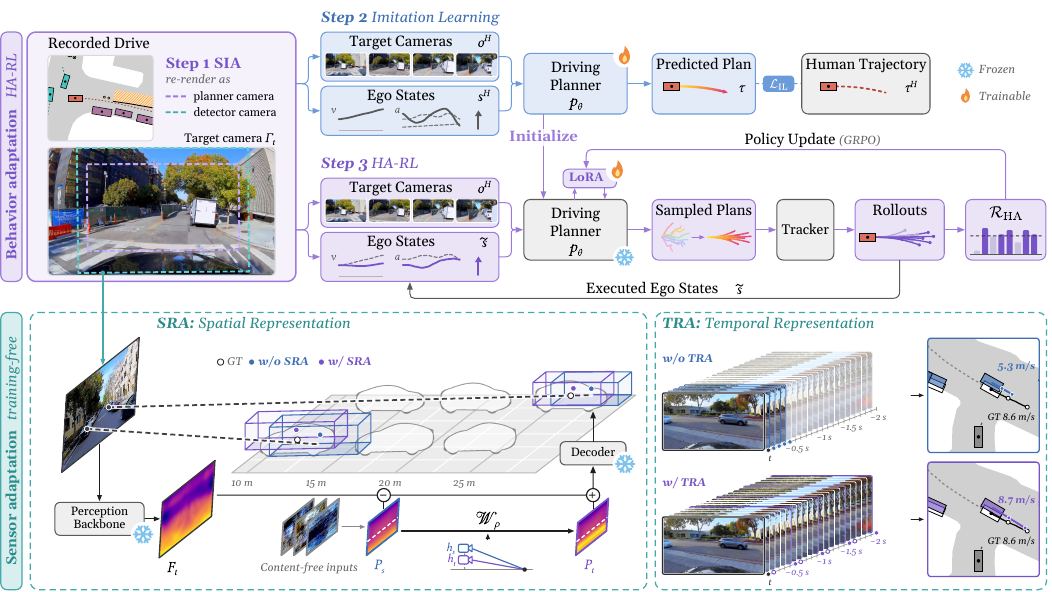}
  \caption{Method of \fw. Step~1, SIA, re-renders the recorded target camera as the source camera of the planner (purple outline) and of the detector (teal outline); both rows start from it. \textbf{Top, behavior adaptation.} Step~2: IL on recorded target driving. Step~3: HA-RL keeps the recorded observation, feeds the planner its executed ego state, and updates a low-rank adapter with the reward $\mathcal{R}_{\mathrm{HA}}$ of a group of executed plans while the planner weights stay frozen. \textbf{Bottom, sensor adaptation} of a released temporal position-embedding detector with frozen weights, after Step~1. SRA: $P_s$ is the mean feature of content-free inputs; replacing it by its re-indexed version $P_t=\mathcal{W}_\rho(P_s)$ moves the boxes toward the ground truth. TRA: at the target frame rate the memory spans only a short time; reading the memory written at the source interval restores this span and the reported speed of a moving vehicle. Deploying a driving model on a new vehicle requires both adaptations: sensor adaptation keeps perception valid on the target rig, and behavior adaptation keeps planning safe under target driving. When the planner acts on detected objects, sensor adaptation is a precondition for behavior adaptation.}
  \label{fig:method}
\end{figure*}

Data and behavior adaptation (top of \cref{fig:method}) aligns the sensor interface and then addresses $\mathcal{D}_t$ and $\mathcal{E}_t$; physical sensor adaptation (bottom) starts from the same alignment and addresses the priors of $\Gamma_s$ left in the perception model.

\textbf{SIA.}
\label{sec:sia}
Sensor-interface alignment (SIA) re-renders the target observations to the source sensor-rig configuration $\Gamma_s$ of the released model (Step~1 in \cref{fig:method}), as detailed, with the architecture-specific corrections, in the supplementary material.
The planner takes the re-rendered cameras in Steps~2 and~3 below, and the perception model takes them as the starting point of sensor adaptation.

\textbf{Target-domain IL.}
\label{sec:il}
The target dataset $\mathcal{D}_t=\{(o_t,s_t,\tau_t^{H})\}_{t=1}^{N}$ is ordinary human driving, where $o_t$ is the sensor observation at time $t$, $s_t$ the ego state and $\tau_t^{H}$ the human-driven future trajectory over a horizon $H$; the surrounding scene is reconstructed for rule-aware evaluation.
Starting from the released parameters $\theta_0$, we fine-tune the planner on $\mathcal{D}_t$ with its original trajectory imitation objective (Step~2 in \cref{fig:method}), yielding $p_{\theta_{\mathrm{IL}}}$.
Direct fine-tuning, however, lets the planner read the driver's action from the ego state: the current acceleration already reveals whether the driver has started to brake, and the planner then slows down from this input rather than from the scene.
We localize this by changing the acceleration input alone with the camera frames fixed, and correct it during fine-tuning: the ego state is computed from past poses only, and the acceleration input is set to zero for half of the training samples and perturbed with noise for the rest.

\textbf{Localizing optimization--execution coupling.}
\label{sec:reward_execution}
IL aligns $p_{\theta}$ with $\mathcal{D}_t$, but the post-training objective can remain coupled to how the source benchmark evaluates trajectories.
At deployment step $k$, the planner predicts $\tau_k=p_{\theta_{\mathrm{IL}}}(o_k,s_k)$, but only the first $\Delta$ seconds are executed before replanning, $s_{k+1}=\mathcal{T}(s_k,\tau_k^{[0,\Delta]})$, where $\mathcal{T}$ denotes vehicle execution.
A common post-training strategy uses the predictive driver model score (PDMS)~\cite{navsim} of the predicted trajectory as a scalar RL reward~\cite{recogdrive,autovla}, which we call the trajectory reward, $\mathcal{R}_{\mathrm{traj}}(\tau_k)=\sum_j w_j r_j(\tau_k)$, with driving-quality terms $r_j$ and weights $w_j$.
We probe this objective by changing the execution interface from trajectory-level evaluation to receding-horizon execution.
The trajectory reward scores the complete trajectory and is insensitive to a speed bias in its first segment $\tau_k^{[0,\Delta]}$, yet this segment is the only part that is executed, and the next plan starts from the state it leaves, so the bias is executed again at every replan and accumulates.
Improving the trajectory reward therefore does not necessarily improve behavior under repeated execution: the post-training objective inherits complete-trajectory evaluation from the source benchmark, whereas the target vehicle repeatedly executes only the initial segment before replanning.

\textbf{HA-RL.}
\label{sec:ha_rl}
A natural correction is generative closed-loop RL in the video-world-model environment used for closed-loop evaluation, where after each executed segment the ego state is updated and the next observation is regenerated by a generative world model~\cite{omnidreams}, so that both become policy-induced, $(o_k^H,s_k^H)\rightarrow(\tilde{o}_k,\tilde{s}_k)$.
Regenerating the camera video for every sampled rollout makes each rollout costly (\cref{sec:exp_planning}) and ties post-training to a world model of the target environment.
The localized mismatch, however, enters through the ego state, while the observation changes little within a rollout of a few seconds that starts from a recorded frame, because the ego pose stays close to the recorded one.
HA-RL therefore intervenes on the execution variable alone, $(o_k^H,s_k^H)\rightarrow(o_k^H,\tilde{s}_k)$: for a recorded sequence $\{(o_k^H,s_k^H)\}_{k=0}^{K}\subset\mathcal{D}_t$, it rolls out (Step~3 in \cref{fig:method})
\begin{equation}
\begin{cases}
    \tau_k
    =
    p_{\theta_{\mathrm{IL}}}(o_k^H,\tilde{s}_k),\\[2pt]
    \tilde{s}_{k+1}
    =
    \mathcal{T}\!\left(
        \tilde{s}_k,\tau_k^{[0,\Delta]}
    \right),
\end{cases}
\qquad
\tilde{s}_0=s_0^H .
\label{eq:ha_rollout}
\end{equation}
Unlike IL, this rollout propagates the consequences of the planner's own decisions; unlike generative closed-loop RL, it keeps the observations of real target driving and requires no video generation.

\textbf{Execution-aware reward.}
Repeated execution of Eq.~\eqref{eq:ha_rollout} produces an executed rollout $\xi_i^{\mathrm{HA}}$, and the execution reward $\mathcal{R}_{\mathrm{exec}}(\xi_i^{\mathrm{HA}})=\sum_j w_j r_j(\xi_i^{\mathrm{HA}})$ applies the same score to this rollout.
Since the mismatch enters as a speed bias in the repeatedly executed first segment, we also anchor this segment to human driving.
Let $\Delta v_i$ be the velocity change in the first $\Delta$ seconds of $\tau_i$ and $[l_b,h_b]$ the empirical range of velocity changes under the corresponding human-driving condition.
We define the human motion anchor
\begin{equation}
\mathcal{R}_{\mathrm{hum}}(\tau_i)
=
I_i
\max\!\left(
0,\,
1-
\frac{d(\Delta v_i,[l_b-1,h_b+1])}
{1\,\mathrm{m/s}}
\right),
\label{eq:human_motion_anchor}
\end{equation}
where $d(x,[a,b])$ denotes the distance from $x$ to the interval $[a,b]$, and $I_i\in\{0,1\}$ indicates whether the executed segment is safe, i.e., free of at-fault collisions and drivable-area violations.
The HA-RL reward is
\begin{equation}
    \mathcal{R}_{\mathrm{HA}}(\tau_i)
    =
    \mathcal{R}_{\mathrm{exec}}(\xi_i^{\mathrm{HA}})
    \left(
        \alpha+(1-\alpha)\mathcal{R}_{\mathrm{hum}}(\tau_i)
    \right),
\label{eq:ha_reward}
\end{equation}
where $\alpha$ weights driving quality under repeated execution against the human speed range of the first segment.

\textbf{Policy update.}
For each recorded frame, we sample a group of plans $\{\tau_i\}$ from the planner; each plan is executed as the first segment of one rollout of Eq.~\eqref{eq:ha_rollout} and receives $\mathcal{R}_{\mathrm{HA}}(\tau_i)$.
The planner is then updated with a group-relative policy-gradient objective~\cite{flowgrpo}, which raises the likelihood of the plans whose reward exceeds the group mean.

\textbf{Sensor adaptation.}
\label{sec:sensor_adaptation}
Sensor adaptation applies the same SIA to the cameras of the perception model and keeps its weights frozen.
Even after this explicit alignment, the released model may retain spatial and temporal priors of the source configuration; we localize them with probes that keep the weights fixed and correct them by SRA and TRA.

\textbf{SRA.}
\label{sec:spatial_alignment}
Spatial-representation alignment (SRA) starts from a probe of spatial coupling: we remove or randomize image content while preserving the sensing geometry, and representation components that remain stable across content changes reveal source-rig structure independent of scene semantics.
Writing an intermediate target-scene representation as $F_t=F_t^{*}+P_s$, with scene-dependent information $F_t^{*}$ and the localized component $P_s$ inherited from the source configuration (a conceptual decomposition, not an assumption that the network is globally additive), we replace only this component with its target-rig counterpart $P_t$:
\begin{equation}
    \bar F_t
    =
    F_t-P_s+P_t
    =
    F_t^{*}+P_t,
\label{eq:spatial_prior_transfer}
\end{equation}
where $P_t$ is derived from the source and target sensor configurations, and the remaining representation and model parameters are unchanged.
The realization of Eq.~\eqref{eq:spatial_prior_transfer} depends on where the architecture encodes its spatial prior.
For a temporal position-embedding detector~\cite{streampetr}, the probe reveals a positional component organized by image row, which appears in the features of camera images as in the mean feature of content-free inputs (\cref{fig:method}, SRA).
We obtain the target component by re-indexing the rows of the source component, $P_t=\mathcal{W}_\rho(P_s)$: $\mathcal{W}_\rho$ stretches the rows below the horizon of the target camera by a factor $\rho$ and keeps the rows above it. A camera at a different height sees the same ground point at a different distance below the horizon, and $\rho$ compensates this difference between the source and target rigs.
For a sparse anchor-sampling detector~\cite{sparse4dv3}, the source configuration is reflected in its 3D reference geometry, including the reference camera height, and we correct this geometry with the target calibration.

\textbf{TRA.}
\label{sec:temporal_alignment}
Temporal-representation alignment (TRA) addresses temporal representations, which can similarly remain coupled to the source interval $\Delta_s$: at a target interval $\Delta_t\neq\Delta_s$, temporal updates learned at $\Delta_s$ take a different physical meaning.
We probe this coupling by varying the update interval with scene content and weights fixed.
The anchor-sampling detector updates the velocity of an instance from its displacement, $v\leftarrow v+\delta/\Delta$, so that at a higher target rate small localization errors become apparent motion; in the position-embedding detector, the temporal prior is implicit in how often the memory is updated.
For both, we update the temporal memory at the source interval while still predicting at every target frame (\cref{fig:method}, TRA): the position-embedding detector reads, at every frame, the memory entries written at multiples of $\Delta_s$ before it, so that the memory covers the same time span as on the source rig, and the anchor-sampling detector writes its memory only at keyframes spaced by $\Delta_s$.

\section{Experiments}
\label{sec:experiments}

\subsection{Setup}
\label{sec:exp_setup}

\textbf{Vehicles and data.}
The planning benchmark is converted from 13.2\,km of downtown driving, scored at 2\,Hz.
It is split by street block into training, validation and test regions, with a buffer around the test region.
Perception is evaluated on three real camera rigs, A, B and C, with one, four and five cameras, recorded at 10\,Hz.
Reviewed LiDAR-based pseudo ground truth is used only for evaluation; no label enters any correction.
For comparison with published methods, detection is also evaluated on the transfer from nuScenes~\cite{nuscenes} to the Lyft~\cite{lyft} and Waymo~\cite{waymo} validation sets in the protocol of CoIn3D~\cite{coin3d}.

The closed loop uses 90 events of 4\,s from the test region: 45 events in which the released planner fails in open loop and 45 events of the same scene types in which it does not.
As in World Engine~\cite{worldengine}, the evaluation focuses on events in which the released model is likely to fail.

\textbf{Models.}
The planners are 13 open end-to-end planners~\cite{transfuser,diffusiondrive,diffusiondrivev2,goalflow,hydramdp,gtrs,drivesuprim,wote,ipad,drivor,recogdrive,autovla}, scored in open loop in the supplementary material, and a diffusion-based world-action planner (SimWAM~\cite{simwam}).
Post-training is evaluated in closed loop on the diffusion-based planner, which takes only the front camera that a generative world model (OmniDreams~\cite{omnidreams}) renders.
HA-RL uses $\alpha = 0.5$ and 100 update steps on four GPUs; the training settings are given in the supplementary material, and RL results are means over three training seeds.
The detectors are a temporal position-embedding detector (StreamPETR~\cite{streampetr}) and a sparse anchor-sampling detector (Sparse4D v3~\cite{sparse4dv3}) with their public nuScenes weights, referred to below as the position-embedding and the anchor-sampling detector; the vehicle stack runs the release of the position-embedding detector in an open-source driving stack (Autoware~\cite{autoware}), trained by its developers with their perception framework~\cite{awml} on their own vehicle rig (T4).

\textbf{Metrics.}
The rule-aware PDMS (RPDMS) extends the NAVSIM v2 score~\cite{navsimv2} with stop-sign compliance, crosswalk yielding and compliance with human traffic control, each applied as a multiplicative term like the traffic-light term;
it is the score used in $\mathcal{R}_{\mathrm{traj}}$ and $\mathcal{R}_{\mathrm{exec}}$.
Planning is scored by open-loop RPDMS on the test region (OL) and by the closed-loop success rate (SR), with the rates of at-fault collisions (Coll.), off-road driving (Off), traffic-rule violations (Rule) and stalls (Stall), in percent of the events; a failed event can fall into several categories; the construction of the benchmark and the terms of RPDMS are described in the supplementary material.
Detection is scored by vehicle mAP, averaged over bird's-eye-view (BEV) center-distance thresholds of 0.5, 1, 2 and 4\,m, within 50\,m of a camera.
We also report the translation, scale and orientation errors of the vehicle matches (mATE, mASE, mAOE) and the nuScenes detection score $\mathrm{NDS}^\ast = \frac{1}{6}\big(3\,\mathrm{mAP} + \sum_{e}(1-\min(1,e))\big)$ over these errors, as in~\cite{coin3d}.
In all tables, the best value of a column is in bold and the second best is underlined, per starting point in \cref{tab:ablation} and per detector in the detection tables; Oracle is excluded.

\textbf{Closed-loop evaluation.}
After HA-RL, we validate the
localized planner in the generative closed-loop environment, where
both the ego state and front-camera observation evolve with the
adapted policy. Each event runs for 4\,s, short enough that the generated frames stay undistorted, with replanning every
0.53\,s. This tests whether the localized correction
remains effective when both evolve jointly.

\subsection{Planning}
\label{sec:exp_planning}

\textbf{Does the planner input need SIA?}
We score the released planner on the whole benchmark with the raw front camera and with the front camera re-rendered into its training camera (SIA, \cref{sec:sia}).
Re-rendering raises RPDMS from 0.623 to 0.691 with the weights unchanged; all planner results below use the re-rendered input.

\textbf{Is the trajectory reward a valid post-training reward?}
From the planner after IL, we post-train with the trajectory reward $\mathcal{R}_{\mathrm{traj}}$ and evaluate in closed loop (\cref{tab:post}).
OL rises from 0.784 to 0.809, the highest in \cref{tab:post}, while SR falls from 84.4\% to 78.1\%, below the released model, and Coll.\ rises from 7.8\% to 12.2\%.
Under the trajectory reward, OL and SR thus move in opposite directions.
Its OL gain over IL comes mostly from the comfort term (0.411 to 0.552), and the progress term falls (0.908 to 0.896).

\textbf{Why does the trajectory reward fail in closed loop?}
We probe each model on fixed camera frames while varying only the input speed, and measure how much faster its first planned segment is than the input speed.
The offset is 0.18\,m/s for the released model, 0.64\,m/s after IL and 0.70\,m/s after post-training with the trajectory reward, which leaves the offset in place while OL rises (\cref{sec:reward_execution}).
The trajectory reward does not register such an offset: when the first segment of the recorded human trajectory is made 1\,m/s faster on 300 training frames, with the path and the later speeds unchanged, its score changes by less than 0.001 on average.
In closed loop, every replan executes this offset again.
After HA-RL, the offset is $-0.07$\,m/s.

\begin{table}[t]
  \centering\footnotesize
  \setlength{\tabcolsep}{3.2pt}\renewcommand{\arraystretch}{0.9}
  \setlength{\aboverulesep}{0.3ex}\setlength{\belowrulesep}{0.4ex}
  \caption{Post-training of the planner on our benchmark. The trajectory reward gives the highest OL but the lowest SR.}
  \label{tab:post}
  \begin{tabular}{@{}l c ccccc@{}}
    \toprule
     & & \multicolumn{5}{c}{Closed loop (\%)} \\
    \cmidrule(l){3-7}
    Training & OL$\uparrow$ & SR$\uparrow$ & Coll.$\downarrow$ & Off$\downarrow$ & Rule$\downarrow$ & Stall$\downarrow$ \\
    \midrule
    Released & 0.698 & 82.2 & 4.4 & 3.3 & \textbf{0.0} & 10.0 \\
    IL & 0.784 & 84.4 & 7.8 & 2.2 & \underline{5.6} & \textbf{0.0} \\
    IL + RL, trajectory reward & \textbf{0.809} & 78.1 & 12.2 & \underline{1.9} & 8.9 & \textbf{0.0} \\
    IL + RL, generative CL & 0.779 & \underline{93.0} & \textbf{1.5} & \underline{1.9} & \textbf{0.0} & 3.7 \\
    IL + HA-RL & \underline{0.785} & \textbf{95.2} & \underline{1.9} & \textbf{1.5} & \textbf{0.0} & \underline{1.5} \\
    \bottomrule
  \end{tabular}
\end{table}

\begin{table}[t]
  \centering\footnotesize
  \setlength{\tabcolsep}{3.2pt}\renewcommand{\arraystretch}{0.9}
  \setlength{\aboverulesep}{0.3ex}\setlength{\belowrulesep}{0.4ex}
  \caption{Ablation of the RL stage: the reward (upper rows) and the starting point (lower rows).}
  \label{tab:ablation}
  \begin{tabular}{@{}ll c ccccc@{}}
    \toprule
     & & & \multicolumn{5}{c}{Closed loop (\%)} \\
    \cmidrule(l){4-8}
    Start & Reward & OL$\uparrow$ & SR$\uparrow$ & Coll.$\downarrow$ & Off$\downarrow$ & Rule$\downarrow$ & Stall$\downarrow$ \\
    \midrule
    IL & $\mathcal{R}_{\mathrm{traj}}$ & \textbf{0.809} & 78.1 & 12.2 & 1.9 & 8.9 & \textbf{0.0} \\
    IL & $\mathcal{R}_{\mathrm{exec}}$ & 0.763 & \underline{94.4} & \underline{2.6} & \textbf{1.1} & \underline{0.4} & \underline{1.5} \\
    IL & $\mathcal{R}_{\mathrm{HA}}$ & \underline{0.785} & \textbf{95.2} & \textbf{1.9} & \underline{1.5} & \textbf{0.0} & \underline{1.5} \\
    \midrule
    Released & $\mathcal{R}_{\mathrm{traj}}$ & -- & 62.6 & 24.8 & 3.3 & 13.0 & \textbf{1.1} \\
    Released & $\mathcal{R}_{\mathrm{HA}}$ & \textbf{0.693} & \textbf{84.8} & \textbf{0.7} & \textbf{0.4} & \textbf{0.0} & 14.1 \\
    \bottomrule
  \end{tabular}
\end{table}

\begin{figure*}[t]
  \centering
  \includegraphics[width=\textwidth]{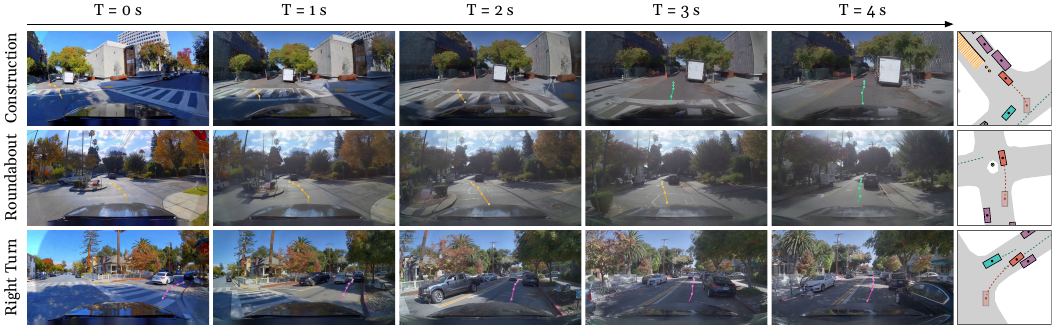}
  \caption{Qualitative results after post-training in three scenes: construction, roundabout, and right turn at an intersection. Each row shows the predicted trajectory on the generated frames of the closed loop and the top view at the end of the event.}
  \label{fig:planqual}
\end{figure*}

\textbf{Does HA-RL improve the closed loop?}
HA-RL raises the SR of the planner from 82.2\% for the released model to 95.2\% (\cref{tab:post}), with Coll.\ falling from 4.4\% to 1.9\% and Stall from 10.0\% to 1.5\%.
IL alone raises OL and removes the stalls, but it raises Coll.\ to 7.8\%; the RL stage with HA-RL removes these collisions at an unchanged OL.
\Cref{fig:planqual} shows this planner in three of the test events.

\textbf{Does the correction need generated observations?}
The three RL rows of \cref{tab:post} start from the same IL model and differ in what follows the planner's execution: nothing under the trajectory reward, the ego state under HA-RL, and both the ego state and the camera observation in the generative closed loop.
Updating the ego state alone raises SR from 78.1\% to 95.2\%, and the generative closed loop reaches 93.0\%.
On the same GPU, a generative rollout takes 17\,s, most of which is video generation, whereas an HA-RL rollout takes 0.9\,s and needs no world model.

\textbf{Which part of the RL stage matters?}
\Cref{tab:ablation} varies the reward and the starting point of the RL stage.
Replacing the trajectory reward by the execution reward alone raises SR from 78.1\% to 94.4\%; adding the human motion anchor keeps OL at the level of IL and reaches 95.2\%.
Without IL, the same RL stage on the released model reaches 62.6\% with the trajectory reward (not scored in open loop) and 84.8\% with HA-RL, with a stall rate of 14.1\%; IL on target driving is therefore needed before the RL stage.

\textbf{Do the post-training stages carry over to a public benchmark?}
We repeat IL and the RL stage for the planner on NAVSIM~\cite{navsim}, with the extended PDMS (EPDMS)~\cite{navsimv2} in the rewards (\cref{tab:navsim}).
OL is EPDMS on navtest, and the closed loop uses 180 events from the test split, selected by how the released model fails when its plans are executed on the recorded frames.
IL raises SR from 53.9\% to 88.3\% and removes all stalls; the RL stage with HA-RL raises it to 90.9\%, with Coll.\ falling from 4.4\% to 2.8\%.

\begin{table}[t]
  \centering\footnotesize
  \setlength{\tabcolsep}{3.2pt}\renewcommand{\arraystretch}{0.9}
  \setlength{\aboverulesep}{0.3ex}\setlength{\belowrulesep}{0.4ex}
  \caption{Post-training on NAVSIM with the stages of \cref{tab:post}.}
  \label{tab:navsim}
  \begin{tabular}{@{}l c ccccc@{}}
    \toprule
     & & \multicolumn{5}{c}{Closed loop (\%)} \\
    \cmidrule(l){3-7}
    Training & OL$\uparrow$ & SR$\uparrow$ & Coll.$\downarrow$ & Off$\downarrow$ & Rule$\downarrow$ & Stall$\downarrow$ \\
    \midrule
    Released & 0.832 & 53.9 & 10.6 & 7.8 & 0.6 & 28.3 \\
    IL & 0.862 & 88.3 & 4.4 & 7.8 & \textbf{0.0} & \textbf{0.0} \\
    IL + RL, trajectory reward & \textbf{0.880} & 89.6 & 3.1 & \underline{7.4} & \underline{0.2} & 0.4 \\
    IL + RL, generative CL & \underline{0.872} & \underline{90.4} & \textbf{2.4} & 7.8 & \textbf{0.0} & \underline{0.2} \\
    IL + HA-RL & 0.865 & \textbf{90.9} & \underline{2.8} & \textbf{5.6} & \textbf{0.0} & 1.5 \\
    \bottomrule
  \end{tabular}
\end{table}

\subsection{Perception}
\label{sec:exp_rig}

\begin{table*}[t]
  \centering\scriptsize
  \setlength{\tabcolsep}{2.9pt}
  \caption{Transfer from nuScenes to Waymo and Lyft in the protocol of CoIn3D. Published methods retrain the detector; R2R keeps the public nuScenes weights frozen.}
  \label{tab:xrig}
  \begin{tabular}{@{}llc|ccccc|ccccc@{}}
    \toprule
     & & & \multicolumn{5}{c|}{nuScenes $\rightarrow$ Waymo} & \multicolumn{5}{c}{nuScenes $\rightarrow$ Lyft} \\
    Detector & Method & Retrain & mAP$\uparrow$ & mATE$\downarrow$ & mASE$\downarrow$ & mAOE$\downarrow$ & \textbf{NDS$^\ast$}$\uparrow$ & mAP$\uparrow$ & mATE$\downarrow$ & mASE$\downarrow$ & mAOE$\downarrow$ & \textbf{NDS$^\ast$}$\uparrow$ \\
    \midrule
    \multirow{8}{*}{Depth-based BEV~\cite{bevdepth}} & \textcolor{gray}{Oracle} & \textcolor{gray}{target} & \textcolor{gray}{0.552} & \textcolor{gray}{0.528} & \textcolor{gray}{0.148} & \textcolor{gray}{0.085} & \textcolor{gray}{0.649} & \textcolor{gray}{0.602} & \textcolor{gray}{0.471} & \textcolor{gray}{0.152} & \textcolor{gray}{0.078} & \textcolor{gray}{0.684} \\
       & Direct Transfer & \textbf{no} & 0.040 & 1.303 & 0.265 & 0.790 & 0.178 & 0.112 & 0.997 & 0.176 & 0.389 & 0.296 \\
       & CAM-Convs~\cite{camconvs} {\scriptsize(CVPR'19)} & yes & 0.045 & 1.301 & 0.253 & 0.773 & 0.185 & 0.145 & 0.999 & 0.173 & 0.368 & 0.316 \\
       & Single-DGOD~\cite{singledgod} {\scriptsize(CVPR'22)} & yes & 0.034 & 1.305 & 0.262 & 0.855 & 0.164 & 0.159 & 0.949 & 0.174 & 0.358 & 0.332 \\
       & DG-BEV~\cite{dgbev} {\scriptsize(CVPR'23)} & yes & 0.297 & 0.822 & \textbf{0.216} & 0.372 & 0.415 & 0.287 & 0.771 & 0.170 & 0.302 & 0.437 \\
       & PD-BEV~\cite{pdbev} {\scriptsize(AAAI'25)} & yes & -- & -- & -- & -- & -- & 0.304 & \underline{0.709} & 0.169 & 0.289 & 0.458 \\
       & UDGA~\cite{udga} {\scriptsize(NeurIPS'24)} & yes & \underline{0.349} & \underline{0.754} & 0.289 & \underline{0.250} & \underline{0.459} & \underline{0.324} & \underline{0.709} & \underline{0.162} & \underline{0.180} & \underline{0.487} \\
       & CoIn3D~\cite{coin3d} {\scriptsize(CVPR'26)} & yes & \textbf{0.381} & \textbf{0.687} & \underline{0.220} & \textbf{0.155} & \textbf{0.513} & \textbf{0.375} & \textbf{0.660} & \textbf{0.161} & \textbf{0.101} & \textbf{0.534} \\
    \midrule
    \multirow{4}{*}{BEV-query~\cite{bevformer}} & Direct Transfer & \textbf{no} & 0.054 & 1.146 & 0.747 & 1.606 & 0.069 & 0.149 & 1.031 & 0.755 & 1.241 & 0.115 \\
       & PD-BEV~\cite{pdbev} {\scriptsize(AAAI'25)} & yes & -- & -- & -- & -- & -- & 0.208 & -- & -- & -- & 0.355 \\
       & CoIn3D~\cite{coin3d} {\scriptsize(CVPR'26)} & yes & \textbf{0.249} & \textbf{0.872} & \underline{0.239} & \textbf{0.365} & \textbf{0.379} & \underline{0.237} & \textbf{0.850} & \underline{0.191} & \underline{0.407} & \underline{0.377} \\
       & \textbf{R2R (ours)} & \textbf{no} & \underline{0.168} & \underline{1.126} & \textbf{0.233} & \underline{0.394} & \underline{0.313} & \textbf{0.270} & \underline{0.913} & \textbf{0.175} & \textbf{0.128} & \textbf{0.432} \\
    \midrule
    \multirow{4}{*}{Position-embedding~\cite{petr}} & Direct Transfer & \textbf{no} & 0.048 & \underline{1.191} & 0.748 & 1.543 & 0.066 & 0.013 & 1.240 & 0.761 & 1.484 & 0.046 \\
       & PD-BEV~\cite{pdbev} {\scriptsize(AAAI'25)} & yes & -- & -- & -- & -- & -- & 0.032 & -- & -- & -- & 0.091 \\
       & CoIn3D~\cite{coin3d} {\scriptsize(CVPR'26)} & yes & \textbf{0.200} & \textbf{1.141} & \underline{0.242} & \underline{0.505} & \underline{0.309} & \textbf{0.332} & \underline{0.880} & \underline{0.170} & \underline{0.210} & \underline{0.456} \\
       & \textbf{R2R (ours)} & \textbf{no} & \underline{0.164} & 1.231 & \textbf{0.214} & \textbf{0.378} & \textbf{0.317} & \underline{0.310} & \textbf{0.830} & \textbf{0.161} & \textbf{0.097} & \textbf{0.474} \\
    \midrule\midrule
    \multirow{2}{*}{Sparse anchor-sampling~\cite{sparse4dv3}} & Direct Transfer & \textbf{no} & 0.171 & 0.928 & \textbf{0.208} & 0.310 & 0.345 & 0.143 & 0.804 & 0.206 & 0.290 & 0.355 \\
       & \textbf{R2R (ours)} & \textbf{no} & \textbf{0.346} & \textbf{0.666} & 0.209 & \textbf{0.154} & \textbf{0.502} & \textbf{0.500} & \textbf{0.520} & \textbf{0.152} & \textbf{0.060} & \textbf{0.628} \\
    \midrule
    \multirow{2}{*}{Temporal position-embedding~\cite{streampetr}} & Direct Transfer & \textbf{no} & 0.070 & 1.239 & 0.286 & 0.580 & 0.224 & 0.065 & 1.036 & 0.291 & 0.613 & 0.215 \\
       & \textbf{R2R (ours)} & \textbf{no} & \textbf{0.232} & \textbf{0.988} & \textbf{0.243} & \textbf{0.212} & \textbf{0.375} & \textbf{0.358} & \textbf{0.754} & \textbf{0.163} & \textbf{0.114} & \textbf{0.507} \\
    \bottomrule
  \end{tabular}
\end{table*}

\textbf{Do the corrections transfer between public datasets?}
\Cref{tab:xrig} applies the corrections to the transfer from nuScenes to Lyft and Waymo, on which published methods retrain the detector for invariance to the camera configuration; cars within 50\,m are scored.
The published rows, including Oracle (trained and tested on the target dataset) and Direct Transfer (the released weights applied with the target calibration), are reported in~\cite{coin3d}, with the tiny BEV-query detector and the single-frame position-embedding detector on VoVNet; a dash marks results not reported.
Our rows keep the public nuScenes weights frozen; R2R applies SIA (re-rendering to the source intrinsics and the source frame) and TRA (2\,Hz chains) to every temporal detector, and SRA as the embedding-height offset for the anchor-sampling detector and $\mathcal{W}_\rho$ for the position-embedding detector, all from calibration.
On Waymo, whose cameras sit 0.75\,m higher than those of nuScenes, record at 10\,Hz and use a reference frame 1.7\,m lower, the corrections raise NDS$^\ast$ from 0.345 to 0.502 for the anchor-sampling detector and from 0.224 to 0.375 for the position-embedding detector.
Without retraining, R2R gives the single-frame position-embedding detector a higher NDS$^\ast$ than CoIn3D obtains for the same detector with retraining on both datasets (0.317 vs.\ 0.309 on Waymo, 0.474 vs.\ 0.456 on Lyft), and the anchor-sampling detector exceeds every retrained method of the table on Lyft (NDS$^\ast$ 0.628 vs.\ 0.534).
On Lyft, whose rig is close to nuScenes, most of the gain comes from placing the principal point on the row of the source camera.
CoIn3D with the depth-based BEV detector, which retrains with synthesized views, remains ahead on Waymo.

\begin{table*}[t]
  \centering\scriptsize
  \setlength{\tabcolsep}{1.1pt}
  \caption{Detection on our three rigs with frozen public nuScenes weights. The alignment steps of \cref{sec:sensor_adaptation} are added in turn, with one fixed setting per detector for all rigs.}
  \label{tab:rig}
  \begin{tabular}{@{}l@{\hspace{2pt}}l@{\hspace{2.1pt}}|@{\hspace{2.1pt}}cccccc@{\hspace{2.1pt}}|@{\hspace{2.1pt}}cccccc@{\hspace{2.1pt}}|@{\hspace{2.1pt}}cccccc@{}}
    \toprule
     & & \multicolumn{6}{c|}{Rig A (1 camera)} & \multicolumn{6}{c|}{Rig B (4 cameras)} & \multicolumn{6}{c}{Rig C (5 cameras)} \\
    Detector & Setting & mAP\kern-0.3pt{\tiny$\uparrow$} & mATE\kern-0.3pt{\tiny$\downarrow$} & mASE\kern-0.3pt{\tiny$\downarrow$} & mAOE\kern-0.3pt{\tiny$\downarrow$} & NDS$^\ast$\kern-0.3pt{\tiny$\uparrow$} & Ped.\kern-0.3pt{\tiny$\uparrow$} & mAP\kern-0.3pt{\tiny$\uparrow$} & mATE\kern-0.3pt{\tiny$\downarrow$} & mASE\kern-0.3pt{\tiny$\downarrow$} & mAOE\kern-0.3pt{\tiny$\downarrow$} & NDS$^\ast$\kern-0.3pt{\tiny$\uparrow$} & Ped.\kern-0.3pt{\tiny$\uparrow$} & mAP\kern-0.3pt{\tiny$\uparrow$} & mATE\kern-0.3pt{\tiny$\downarrow$} & mASE\kern-0.3pt{\tiny$\downarrow$} & mAOE\kern-0.3pt{\tiny$\downarrow$} & NDS$^\ast$\kern-0.3pt{\tiny$\uparrow$} & Ped.\kern-0.3pt{\tiny$\uparrow$} \\
    \midrule
    \multirow{4}{*}{\begin{tabular}{@{}l@{}}Temporal\\position-\\embedding~\cite{streampetr}\end{tabular}} & Released & 0.014 & 1.445 & 0.262 & \underline{0.161} & 0.270 & 0.000 & 0.059 & 1.217 & 0.290 & 0.314 & 0.262 & 0.152 & 0.041 & 1.218 & 0.285 & 0.293 & 0.257 & 0.112 \\
     & + SIA & 0.211 & 1.017 & 0.232 & 0.296 & 0.351 & 0.026 & 0.187 & 1.080 & \underline{0.244} & 0.245 & 0.345 & \underline{0.322} & 0.200 & 1.060 & 0.263 & 0.283 & 0.343 & \underline{0.251} \\
     & + SRA & \underline{0.243} & \underline{0.960} & \underline{0.227} & 0.282 & \underline{0.377} & \underline{0.053} & \underline{0.192} & \underline{1.056} & 0.249 & \underline{0.241} & \underline{0.348} & 0.314 & \underline{0.207} & \underline{1.031} & \underline{0.253} & \underline{0.264} & \underline{0.351} & 0.216 \\
     & + TRA (\textbf{R2R}) & \textbf{0.475} & \textbf{0.852} & \textbf{0.154} & \textbf{0.111} & \textbf{0.551} & \textbf{0.315} & \textbf{0.403} & \textbf{0.895} & \textbf{0.237} & \textbf{0.196} & \textbf{0.480} & \textbf{0.489} & \textbf{0.420} & \textbf{0.898} & \textbf{0.220} & \textbf{0.190} & \textbf{0.492} & \textbf{0.297} \\
    \midrule
    \multirow{4}{*}{\begin{tabular}{@{}l@{}}Sparse\\anchor-\\sampling~\cite{sparse4dv3}\end{tabular}} & Released & 0.392 & 0.960 & \underline{0.126} & 0.181 & 0.485 & 0.124 & 0.334 & 0.913 & 0.237 & 0.175 & 0.446 & 0.667 & 0.327 & 0.931 & 0.226 & 0.206 & 0.436 & \textbf{0.316} \\
     & + SIA & \underline{0.430} & 0.916 & 0.129 & 0.170 & 0.512 & 0.159 & 0.343 & 0.892 & 0.234 & \textbf{0.163} & 0.457 & 0.738 & 0.334 & 0.903 & 0.224 & 0.193 & 0.447 & 0.292 \\
     & + SRA & \textbf{0.467} & \textbf{0.847} & 0.133 & \underline{0.159} & \underline{0.543} & \underline{0.182} & \underline{0.349} & \underline{0.870} & \underline{0.233} & \underline{0.164} & \underline{0.463} & \textbf{0.828} & \underline{0.356} & \underline{0.874} & \textbf{0.217} & \underline{0.189} & \underline{0.465} & \underline{0.315} \\
     & + TRA (\textbf{R2R}) & \textbf{0.467} & \underline{0.864} & \textbf{0.123} & \textbf{0.100} & \textbf{0.552} & \textbf{0.248} & \textbf{0.491} & \textbf{0.744} & \textbf{0.214} & 0.183 & \textbf{0.556} & \underline{0.785} & \textbf{0.451} & \textbf{0.825} & \underline{0.218} & \textbf{0.153} & \textbf{0.526} & 0.296 \\
    \bottomrule
  \end{tabular}
\end{table*}

\textbf{Do the corrections transfer to our rigs?}
\Cref{tab:rig} adds the three steps of \cref{sec:sensor_adaptation} in turn, with one fixed setting per detector and frozen public nuScenes weights on our three rigs.
For the position-embedding detector, SIA re-renders the cameras with the source intrinsics, SRA applies $\mathcal{W}_\rho$ with $\rho$ from the camera heights, and TRA writes the memory at the source interval as 2\,Hz chains; for the anchor-sampling detector, SIA places the model frame as on the source rig, SRA offsets the embedding height, and TRA uses a keyframe memory.
Without labels, the full set raises vehicle mAP on every rig, to 0.40--0.48 for the position-embedding and 0.45--0.49 for the anchor-sampling detector; on rig C, the former rises from 0.041 to 0.420.
Every step raises the vehicle mAP of both detectors on most rigs.

\textbf{Where does the training rig reside?}
Probes with fixed image content isolate each prior (measurements in the supplementary material).
The content-free field of the position-embedding detector is organized by image row; re-indexing its rows changes the predicted range.
For the anchor-sampling detector, an embedding-height offset alone restores the near-range distance, and the opposite offset cancels a lowered reference frame, so the camera height resides in the embedding.
The frame interval shows as a speed given to parked vehicles, which the source interval removes.

\textbf{Which correction helps on which rig?}
Both detectors come from the nuScenes rig, whose cameras have a narrower field of view than ours and sit 0.2--0.3\,m lower, yet the corrections that matter differ.
The position-embedding detector does not sample image features through the calibrated projection: SIA gives its first large gain (0.041 to 0.200 on rig C), TRA adds as much again (0.207 to 0.420), and SRA adds most on rig A (0.211 to 0.243).
The anchor-sampling detector projects its anchors with the target calibration and starts higher: SRA adds most on rig A (0.430 to 0.467), and TRA on rigs B and C (0.349 to 0.491 and 0.356 to 0.451).
Re-rendering to the narrower nuScenes intrinsics discards part of our view and lowers the anchor-sampling detector on rig A, so its SIA only places the model frame.

\textbf{Does the R2R detector run unchanged in the vehicle stack?}
The vehicle stack runs its own release of the position-embedding detector.
We package the three steps for the unmodified node of the driving stack: a front end re-renders the cameras, $\mathcal{W}_\rho$ is a constant added to the output of the exported backbone graph, and five node instances form the 2\,Hz chains.
The calibration of its training rig is not public, so $\rho$ is set on rig A and kept on rig C.
Replayed offline on the five-camera drive, the node reaches a vehicle mAP of 0.477, against 0.204 for the release.

\section{Conclusion}
\label{sec:conclu}
In this work, we introduced R2R, a methodology for deploying released driving models in new environments and on new vehicles with minimal effort.
Our results show that behavior adaptation and sensor adaptation improve both planning and perception.
This work demonstrates that a low-cost path to real-vehicle deployment is feasible.
Looking forward, we see opportunities in automating the search for source-system coupling and in extending R2R to new environments, new sensor configurations and more open-source model architectures.

{
    \small
    \bibliographystyle{ieeenat_fullname}
    \bibliography{main}
}

\ifdefined\oldtwocolumn\let\twocolumn\oldtwocolumn\fi  %
\makeatletter  %
\let\arx@setcounter\setcounter
\def\setcounter#1#2{\def\arx@tmp{#1}\def\arx@page{page}\ifx\arx@tmp\arx@page\else\arx@setcounter{#1}{#2}\fi}
\makeatother
\clearpage
\setcounter{page}{1}
\maketitlesupplementary
\appendix

\section{Localizing the Couplings}
\label{sec:supp_probes}

Each correction of R2R follows a probe that changes one input of the released model while the scene content and the weights stay fixed.
\Cref{tab:probes} lists the probes behind the corrections of the main paper and what they measure.
In each case the probe isolates a single input through which the source system enters, and the correction acts on that input only.

\begin{table*}[t]
  \centering\footnotesize
  \caption{Probes that localize each coupling. Every probe changes one input while the scene content and the model weights stay fixed. Range ratio: median of predicted over true distance for matched vehicles at 0--10\,m on rig A, after SIA.}
  \label{tab:probes}
  \begin{tabular}{@{}p{2.8cm}p{4.1cm}p{5.6cm}p{3.3cm}@{}}
    \toprule
    Model and prior & Probe & Measurement & Correction \\
    \midrule
    Planner, ego-state input & Acceleration input changed, camera frames fixed & The plan follows the input acceleration instead of the scene & IL with past-pose ego state and masked acceleration \\
    Planner, post-training objective & Input speed varied on fixed frames; first segment of recorded trajectories made 1\,m/s faster & First-segment speed offset of 0.18, 0.64 and 0.70\,m/s for the released, IL and trajectory-reward models; trajectory score changes by less than 0.001 & HA-RL (offset $-0.07$\,m/s) \\
    \midrule
    Position-embedding detector, camera height & Content-free inputs; rows of the positional component re-indexed by $\mathcal{W}_\rho$ & Component organized by image row; range ratio from 0.92 to 0.95 & SRA: $\mathcal{W}_\rho$ \\
    Driving-stack release, frame interval & Memory written every frame with time stamps scaled to 0.5\,s, or written every 0.5\,s & Vehicle mAP 0.420 with writes at every frame, 0.409 with scaled time stamps, 0.481 with writes every 0.5\,s (\cref{tab:keyframe}) & TRA: memory written at the source interval \\
    Anchor-sampling detector, camera height & Height input of the anchor embedding offset; geometry, projection and box regression fixed & Range ratio from 0.94 to 0.98; the opposite offset cancels a lowered reference frame & SRA: embedding-height offset \\
    Anchor-sampling detector, frame interval & Update interval of 0.1 or 0.5\,s; time stamps changed alone & Parked vehicles receive a median speed of 0.41\,m/s at 10\,Hz, which 2\,Hz updates remove; 2\,Hz input stamped 0.1\,s apart brings it back (1.90\,m/s) & TRA: keyframe memory \\
    \bottomrule
  \end{tabular}
\end{table*}

\section{Benchmark Construction}
\label{sec:supp_bench}

\textbf{Recorded drive.}
The benchmark is built from one ordinary human drive.
The released planner is scored in open loop on every frame of the drive, which locates the frames on which it fails.
The same recording supplies the demonstrations for IL and the starting observations of the closed loop.
It covers 51\,min and 13.2\,km in 9 segments, recorded with five cameras, a 128-beam LiDAR and an INS.
Poses come from LiDAR odometry with loop closures anchored to the INS, and overlapping segments agree to a median of 0.06\,m.
The scenes contain 10 signalized intersections, 50 stop signs, 2 locations with a traffic controller and 652 frames in construction zones.
Street blocks are assigned to a training, a validation and a test region.

\textbf{Ground-truth objects.}
Object tracks come from LiDAR detection and tracking and are reviewed by a vision-language agent at 5\,Hz.
Because the human drive is free of incidents, scorer violations of the human trajectory are used to find and correct errors in the poses, tracks and map.

\textbf{Lanes and traffic control.}
Lanes come from a commercial HD map aligned to our LiDAR.
On the 2.6\,km of streets that the map does not cover, lanes are built from OpenStreetMap topology, LiDAR curbs and reflective paint.
Traffic signals and stop signs come from OpenStreetMap and are checked on the camera images; the signal state of each frame is read from the cameras.

\textbf{Rule-aware score.}
The benchmark adds three traffic rules to the extended PDM score~\cite{navsimv2} as multiplicative terms:
\begin{equation*}
\begin{aligned}
\mathrm{RPDMS} = {} & \mathrm{NC}\cdot\mathrm{DAC}\cdot\mathrm{DDC\text{-}R}\cdot\mathrm{TLC} \\
& \cdot\,\mathrm{SSC}\cdot\mathrm{CWY}\cdot\mathrm{HCC} \\
& \cdot\,\frac{5\,\mathrm{EP} + 5\,\mathrm{TTC} + 2\,\mathrm{LK} + 2\,\mathrm{HC} + 2\,\mathrm{EC}}{16}.
\end{aligned}
\end{equation*}
SSC scores the lowest speed within 10\,m before a stop line at which the vehicle has not yet stopped, CWY scores the distance to pedestrians on a crosswalk that the ego path crosses, and HCC scores whether the ego follows a human traffic controller.
DDC-R judges the driving direction from the lane heading, and EP is normalized by PDM-Closed; the remaining terms are those of the extended PDM score~\cite{navsimv2}.

\textbf{Released planners on the benchmark.}
\Cref{tab:bench} scores 13 open-source planners and SimWAM on the 1736 frames of the test region, each with the vehicle-state inputs it was trained with.
All of them score below the human driver and PDM-Closed.
The most frequent failure is driving into a stopped lead vehicle when a queue starts to move, which 9 of the 13 planners do on at least half of the frames in which the vehicle waits in a queue.
With the ego state built from past poses only, as in post-training, SimWAM scores 0.698 (\cref{tab:post}).

\begin{table*}[t]
  \centering\small
  \setlength{\tabcolsep}{3.6pt}
  \caption{Open-loop RPDMS and its terms on the test region. NC to HCC multiply the score; EP to EC enter the weighted average.}
  \label{tab:bench}
  \begin{tabular}{l c ccccccc ccccc}
    \toprule
    Planner & RPDMS & NC & DAC & DDC-R & TLC & SSC & CWY & HCC & EP & TTC & LK & HC & EC \\
    \midrule
    Human (recorded) & 0.913 & 1.000 & 1.000 & 1.000 & 1.000 & 0.835 & 1.000 & 1.000 & 0.823 & 1.000 & 1.000 & 0.849 & 0.964 \\
    PDM-Closed & 0.880 & 0.980 & 0.994 & 1.000 & 1.000 & 0.901 & 0.995 & 1.000 & 0.996 & 0.975 & 0.987 & 0.809 & 0.518 \\
    \midrule
    DrivoR~\cite{drivor} & 0.782 & 0.945 & 0.980 & 0.986 & 0.984 & 0.794 & 0.994 & 1.000 & 0.889 & 0.944 & 0.849 & 0.846 & 0.777 \\
    Hydra-MDP~\cite{hydramdp} & 0.767 & 0.922 & 0.992 & 0.993 & 0.984 & 0.755 & 0.994 & 1.000 & 0.872 & 0.927 & 0.873 & 0.849 & 0.792 \\
    GTRS-Dense~\cite{gtrs} & 0.765 & 0.968 & 0.988 & 0.988 & 0.988 & 0.710 & 0.994 & 1.000 & 0.813 & 0.968 & 0.829 & 0.838 & 0.617 \\
    Latent TransFuser~\cite{transfuser} & 0.759 & 0.918 & 0.969 & 0.983 & 0.990 & 0.734 & 0.995 & 1.000 & 0.858 & 0.918 & 0.893 & 0.849 & 0.896 \\
    GoalFlow~\cite{goalflow} & 0.754 & 0.939 & 0.973 & 0.983 & 0.988 & 0.798 & 0.994 & 1.000 & 0.858 & 0.942 & 0.874 & 0.848 & 0.549 \\
    DiffusionDrive~\cite{diffusiondrive} & 0.748 & 0.905 & 0.972 & 0.981 & 0.986 & 0.769 & 0.994 & 1.000 & 0.865 & 0.906 & 0.861 & 0.849 & 0.881 \\
    DriveSuprim~\cite{drivesuprim} & 0.748 & 0.901 & 0.991 & 0.990 & 0.983 & 0.752 & 0.994 & 1.000 & 0.904 & 0.906 & 0.892 & 0.850 & 0.739 \\
    TransFuser~\cite{transfuser} & 0.738 & 0.914 & 0.957 & 0.972 & 0.987 & 0.774 & 0.995 & 1.000 & 0.852 & 0.922 & 0.868 & 0.847 & 0.913 \\
    iPad~\cite{ipad} & 0.736 & 0.927 & 0.990 & 0.986 & 0.984 & 0.789 & 0.993 & 1.000 & 0.845 & 0.935 & 0.885 & 0.839 & 0.577 \\
    WoTE~\cite{wote} & 0.730 & 0.916 & 0.969 & 0.979 & 0.986 & 0.796 & 0.995 & 1.000 & 0.831 & 0.919 & 0.838 & 0.844 & 0.753 \\
    SimWAM~\cite{simwam} & 0.705 & 0.926 & 0.988 & 0.993 & 0.984 & 0.799 & 0.994 & 1.000 & 0.888 & 0.927 & 0.853 & 0.814 & 0.184 \\
    ReCogDrive~\cite{recogdrive} & 0.698 & 0.896 & 0.975 & 0.985 & 0.984 & 0.831 & 0.994 & 1.000 & 0.891 & 0.891 & 0.828 & 0.850 & 0.407 \\
    AutoVLA~\cite{autovla} & 0.680 & 0.880 & 0.927 & 0.967 & 0.987 & 0.752 & 0.994 & 1.000 & 0.853 & 0.884 & 0.880 & 0.843 & 0.707 \\
    DiffusionDriveV2~\cite{diffusiondrivev2} & 0.646 & 0.899 & 0.972 & 0.977 & 0.983 & 0.796 & 0.994 & 1.000 & 0.838 & 0.899 & 0.814 & 0.721 & 0.230 \\
    \bottomrule
  \end{tabular}
\end{table*}

\section{Post-Training Settings}
\label{sec:supp_post}

\textbf{Training of the planner.}
IL fine-tunes the full planner with its official objective for 1000 steps (learning rate $2\cdot10^{-5}$, batch 8).
The ego-state input is built from past poses, and the acceleration input is set to zero with probability 0.5 and otherwise perturbed by $\mathcal{N}(0, 0.3)$, so that the planner cannot read the human action from the current acceleration.
HA-RL trains a low-rank adapter of rank 16 on the action branch with Flow-GRPO~\cite{flowgrpo}: groups of 8 plans per scene, exploration noise 0.03 from a fixed initial noise, PPO clipping 0.005 and learning rate $5\cdot10^{-5}$, for 100 update steps on four GPUs, each collecting 32 scenes that are reused for four update steps.

\textbf{Post-training on NAVSIM.}
On NAVSIM, IL uses 21{,}826 frames of the navtrain logs, and RL uses 1{,}512 of them, half of which are frames on which the IL model collides, leaves the drivable area or stalls in the rollout of \cref{eq:ha_rollout}.
The rewards use EPDMS, and the remaining settings are those above.

\section{Closed-Loop Simulation with the World Model}
\label{sec:supp_cl}

\begin{figure*}[t]
  \centering
  \includegraphics[width=\textwidth]{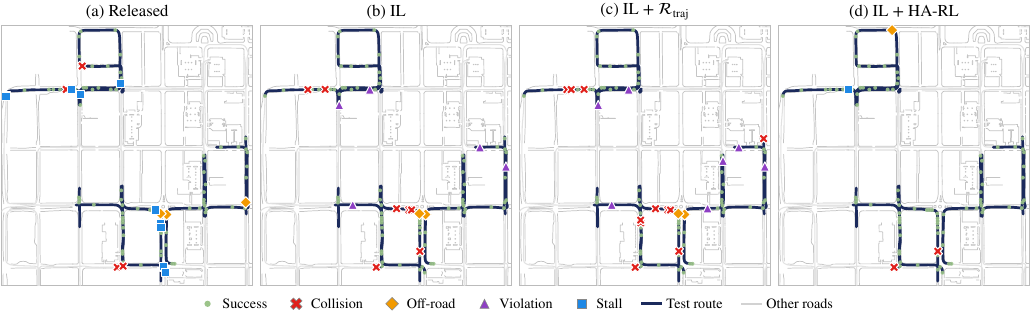}
  \caption{Outcomes of the closed-loop test events on the map. (a) Released. (b) IL. (c) IL and RL with the trajectory reward $\mathcal{R}_{\mathrm{traj}}$. (d) IL and HA-RL. (c) and (d) show one training seed. Markers are placed at the start of each event.}
  \label{fig:clmap}
\end{figure*}

\begin{table*}[t]
  \centering\footnotesize
  \setlength{\tabcolsep}{4pt}
  \caption{Detection in 16\,s closed-loop planning on the benchmark. Vehicle mAP of the detections given to the planner, and success rate (SR), at-fault collisions (Coll.), off-road driving (Off) and rule violations (Rule) in percent of the events.}
  \label{tab:p2p}
  \begin{tabular}{ll c cccc c cccc}
    \toprule
     & & & \multicolumn{4}{c}{PDM-Closed} & & \multicolumn{4}{c}{Diffusion Planner} \\
    \cmidrule{4-7}\cmidrule{9-12}
    Detector & Setting & mAP$\uparrow$ & SR$\uparrow$ & Coll.$\downarrow$ & Off$\downarrow$ & Rule$\downarrow$ & & SR$\uparrow$ & Coll.$\downarrow$ & Off$\downarrow$ & Rule$\downarrow$ \\
    \midrule
    -- & Ground truth & -- & 89.7 & 6.9 & 2.5 & 1.1 &  & 56.8 & 29.5 & 13.6 & 8.1 \\
    \midrule
    \multirow{2}{*}{Temporal position-embedding~\cite{streampetr}} & Released & 0.029 & 71.4 & 25.8 & \textbf{2.2} & 2.2 &  & 48.0 & 38.3 & 14.1 & 14.7 \\
     & R2R & \textbf{0.450} & \textbf{79.9} & \textbf{15.2} & 3.3 & \textbf{1.4} &  & \textbf{60.7} & \textbf{25.8} & \textbf{12.1} & \textbf{9.5}  \\
    \midrule
    \multirow{2}{*}{Sparse anchor-sampling~\cite{sparse4dv3}} & Released & 0.346 & 79.9 & 17.0 & \textbf{2.8} & \textbf{0.9} &  & 56.1 & 31.0 & \textbf{11.8} & \textbf{10.5}  \\
     & R2R & \textbf{0.446} & \textbf{80.5} & \textbf{15.2} & 3.0 & 1.6 &  & \textbf{58.0} & \textbf{27.3} & 12.1 & 11.8 \\
    \midrule
    \multirow{2}{*}{Driving-stack release~\cite{autoware}} & Released & 0.204 & 76.3 & 19.9 & \textbf{2.7} & \textbf{1.1} &  & \textbf{59.4} & \textbf{26.9} & 12.2 & \textbf{10.1}  \\
     & R2R & \textbf{0.405} & \textbf{82.4} & \textbf{12.7} & 3.5 & 1.5 &  & 58.9 & 27.2 & \textbf{11.9} & 11.5 \\
    \bottomrule
  \end{tabular}
\end{table*}

\textbf{Simulation setup.}
The closed-loop results of post-training run the planner inside a generative world model.
The world model renders the front camera at 1280$\times$704, and the image is re-rendered to the planner's camera.
A 30\,Hz tracker executes the plan, and the planner replans every 16 generated frames (0.53\,s).
Other road users follow their recorded trajectories, and each simulation starts 1.2\,s before the event.
\Cref{fig:clmap} shows the outcome of every test event on the map.

\textbf{Fidelity of the simulation.}
Two checks show that the simulation itself does not cause failures.
Driven in the simulation, the recorded human trajectory completes all 90 events.
On real frames, the images that the simulation produces lead to almost the same plans as the original images: the 4\,s endpoints differ by a median of 0.15\,m, and the go or stop decision agrees on 99.0\% of frames.

\textbf{Recorded frames in the HA rollout.}
The HA rollout keeps the recorded frames and replaces only the ego state, which makes it cheap enough to score every sampled plan.
\Cref{tab:harank} checks that it orders policies as the world-model closed loop does.
The released policy is compared with two policies post-trained from it with trajectory rewards, with and without the comfort term, whose first planned segment runs faster than the input speed.
On each test event, the difference of a score between two policies is compared with their difference in closed-loop success.
The HA rollout score follows the closed loop on most events where the outcomes differ, and the trajectory score does not.

\begin{table}[t]
  \centering\footnotesize
  \caption{Agreement of two scores with the world-model closed loop on the 90 test events, for the released policy against two policies post-trained from it with trajectory rewards. Sign: share of the events with different closed-loop outcomes on which the score difference has the sign of the success difference. $\tau$: Kendall rank correlation of the two differences over all events.}
  \label{tab:harank}
  \begin{tabular}{l cc cc}
    \toprule
     & \multicolumn{2}{c}{Without comfort term} & \multicolumn{2}{c}{With comfort term} \\
    \cmidrule(lr){2-3}\cmidrule(lr){4-5}
    Score & Sign & $\tau$ & Sign & $\tau$ \\
    \midrule
    Trajectory score & 0.43 & $-0.04$ & 0.49 & $-0.05$ \\
    HA rollout score & 0.73 & 0.41 & 0.69 & 0.35 \\
    \bottomrule
  \end{tabular}
\end{table}

\textbf{Definition of a stall.}
An event counts as stalled when the vehicle stops early: its progress is below half of the human's and it comes to rest (end speed below 0.5\,m/s, or below 0.3\,m/s for at least 1.5\,s).
A stop with a reason does not count: the human also stopped, the vehicle waits within 15\,m before a stop line at which it has not yet stopped (6\,m for a stop sign with nothing in between), or another road user blocks the path ahead.
Stopping beyond a stop line, inside an intersection or on a crosswalk always counts as a stall.
The rule agrees with manual inspection on 47 of 48 events of the released model.

\section{Detectors on New Camera Rigs}
\label{sec:supp_perception}

\textbf{Detection metrics.}
Perception is evaluated against the same reviewed LiDAR tracks.
A box is evaluated if its center lies in the native view of at least one real camera, between 1\,m and 50\,m in depth from that camera.
Vehicles comprise cars, trucks and buses.
Detections are matched greedily by score at BEV center distances of 0.5, 1, 2 and 4\,m, AP follows the nuScenes definition, and mAP averages the four thresholds.
The evaluation sets contain 1990, 8545 and 9676 annotated vehicle boxes on rigs A, B and C, and 363, 172 and 293 annotated pedestrian boxes.

\textbf{Camera rigs.}
The released detectors were trained on rigs that differ from ours in camera height, pitch and field of view (\cref{tab:rigs}).
The T4 rig of the driving-stack release tilts its side cameras about 18\textdegree{} down, while our cameras are close to level, and the nuScenes rig mounts its cameras about 0.2\,m lower than ours.

\begin{table}[t]
  \setlength{\tabcolsep}{3pt}
  \renewcommand{\arraystretch}{1.3}
  \caption{Camera rigs. Height above the ground, pitch below the horizontal (negative: upward) and horizontal field of view of the native image, as ranges over the cameras of a rig. nuScenes: source of the position-embedding and anchor-sampling detectors; T4: source of the driving-stack release, front / side cameras.}
  \label{tab:rigs}
  \centering\footnotesize
  \begin{tabular}{|l|c|c|c|c|}
    \hline
    \textbf{Rig} & \textbf{Cameras} & \textbf{Height (m)} & \textbf{Pitch (\textdegree)} & \textbf{HFOV (\textdegree)} \\
    \hline
    nuScenes~\cite{nuscenes} & 6 & 1.50--1.59 & $-1$--1 & 70--110 \\
    T4~\cite{awml} & 5 & 1.95 / 1.72 & $-0.9$ / 17.8--19.6 & 70 \\
    \hline
    Ours, A & 1 & 1.78 & 3.1 & 89.5 \\
    Ours, B & 4 & 1.74--1.82 & 0.5--2.6 & 89.5 \\
    Ours, C & 5 & 1.73--1.78 & 0.5--3.9 & 85.9--89.1 \\
    \hline
  \end{tabular}
\end{table}

\textbf{Settings of SIA, SRA and TRA.}
For the position-embedding detector, SIA re-renders each camera with the intrinsics of the nuScenes camera at the same mounting position, SRA applies $\mathcal{W}_\rho$ (\cref{eq:spatial_prior_transfer}) with $\rho$ the ratio of the source and target camera heights, and TRA runs five copies of the temporal memory that start 0.1\,s apart, so that each copy is updated every 0.5\,s as in training.
For the anchor-sampling detector, SIA places the reference frame of the model where the nuScenes rig places it relative to the front camera, SRA adds the camera-height difference to the height input of the anchor embedding, and TRA keeps one memory that is read every frame and written every 0.5\,s.
For both detectors with public nuScenes weights, every parameter follows from the calibrations of the source and target rigs.
The training rig of the driving-stack release is not public: its SIA renders each camera as a virtual pinhole with the field of view of the T4 cameras, turns the side cameras to the pitch of the T4 slots and places the model frame as on the T4 rig, its SRA adds the camera-height difference to the image-side position encoding and applies $\mathcal{W}_\rho$ with $\rho$ set on rig A, and its TRA writes the memory at 2\,Hz.

\textbf{Keyframe memory in the vehicle stack.}
The main paper forms the 2\,Hz chains of the driving-stack release with five node instances.
The network runs once per frame in either case, so the five instances add no computation, only four further copies of the model in GPU memory.
A single memory that is read at every frame and written every 0.5\,s removes these copies and is at least as accurate (\cref{tab:keyframe}): on rig C it reaches a vehicle mAP of 0.481, against 0.477 for the five instances.
This single-instance keyframe memory runs in the driving stack on our vehicle.
The same table separates the cause: scaling only the time stamps or only extending the memory span recovers little of the gain, so the write interval of the memory carries the temporal prior.

\begin{table}[t]
  \centering\footnotesize
  \caption{Temporal memory of the driving-stack release on rig C, in the configuration of the main paper. Every variant outputs detections at every frame (10\,Hz).}
  \label{tab:keyframe}
  \begin{tabular}{lcc}
    \toprule
    Memory & Vehicle mAP & Ped.\ mAP \\
    \midrule
    Read and written every frame & 0.420 & 0.171 \\
    \quad time stamps scaled to 0.5\,s & 0.409 & 0.203 \\
    \quad newest block every frame, 2\,s span & 0.437 & 0.230 \\
    Five instances, each written every 0.5\,s & 0.477 & 0.231 \\
    One instance, written every 0.5\,s & \textbf{0.481} & \textbf{0.271} \\
    \bottomrule
  \end{tabular}
\end{table}

\begin{figure}[t]
  \centering
  \includegraphics[width=\linewidth]{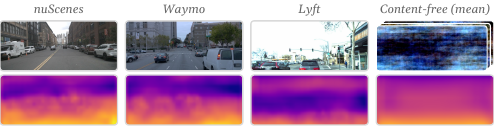}
  \caption{Camera images from nuScenes, Waymo and Lyft and content-free inputs (top), and their backbone features along the first positional direction for the detector of \cref{fig:method} (bottom; same rows and color scale). The last map is the mean over the content-free inputs, $P_s$.}
  \label{fig:posrows}
\end{figure}

\textbf{Position-only component of the image features.}
Content-free inputs are real camera images with randomized Fourier phase, which keeps their power spectrum and removes their layout, together with Gaussian-noise images; $P_s$ is their mean feature.
They expose a component of the backbone features that depends only on the pixel position and mainly on the image row.
\Cref{fig:posrows} shows that the same row pattern appears in the features of real camera images from nuScenes, Waymo and Lyft.

\section{Effect of Detection on Closed-Loop Planning}
\label{sec:supp_p2p}

This section tests whether the detection gains of R2R carry over to driving.
We pass the detections of each detector, released and with R2R, to two planners that take detected objects as input, PDM-Closed and Diffusion Planner, and drive 16\,s in closed loop from events that start every 1\,s on the lanes of the HD map in all three regions (\cref{tab:p2p}).
The planner replans every 0.5\,s from the state the vehicle has reached, other road users follow their recorded trajectories, and the executed trajectory is scored in four consecutive 4\,s segments, each on the scene at its start.
An event succeeds if no segment has an at-fault collision, off-road driving, a rule violation or a stall.
Detections above a fixed score threshold per detector and class are given to the planner, and their vehicle mAP is computed on the frames of the events; the ground-truth objects inside the camera views serve as a reference input.
With ground-truth objects, PDM-Closed succeeds on 89.7\% of the events and Diffusion Planner on 56.8\%, so the success of the latter is bounded by the planner itself.
With R2R, PDM-Closed succeeds more often with all three detectors, mainly through fewer collisions, and Diffusion Planner with the two detectors trained on nuScenes.

\end{document}